\documentclass[conference]{IEEEtran}
\IEEEoverridecommandlockouts
\usepackage{cite}
\usepackage{amsmath,amssymb,amsfonts, nccmath}
\usepackage{algorithm}
\usepackage{algpseudocode}
\usepackage{graphicx}
\usepackage{textcomp}
\usepackage{xcolor}
\usepackage{bbm}
\usepackage{bm}

\usepackage{url}
\date{\vspace{-5ex}}
\usepackage{subcaption}
\usepackage{pdfpages}
\usepackage{adjustbox}
\usepackage{nicefrac} 
\usepackage{enumitem}

\usepackage{parskip}
\def\BibTeX{{\rm B\kern-.05em{\sc i\kern-.025em b}\kern-.08em
    T\kern-.1667em\lower.7ex\hbox{E}\kern-.125emX}}

\begin{document}

\title{CAMEO: Curiosity Augmented Metropolis for Exploratory Optimal Policies} 

\author{\IEEEauthorblockN{Simo Alami Chehboune}
\IEEEauthorblockA{
\textit{Ecole Polytechnique/IRT SystemX}\\
Palaiseau, France \\
mohamed.alami-chehboune@polytechnique.edu}
\and
\IEEEauthorblockN{Fernando Llorente}
\IEEEauthorblockA{\textit{Universidad Carlos III de Madrid}\\
Leganés, Spain \\
felloren@est-econ.uc3m.es}
\and
\IEEEauthorblockN{Rim Kaddah}
\IEEEauthorblockA{\textit{IRT SystemX}\\
Palaiseau, France \\
rim.kaddah@irt-systemx.fr}
\and
\IEEEauthorblockN{Luca Martino}
\IEEEauthorblockA{
\textit{Universidad Rey Juan Carlos}\\
Fuenlabrada, Spain \\
luca.martino@urjc.es}
\and
\IEEEauthorblockN{Jesse Read}
\IEEEauthorblockA{\textit{LIX, Ecole Polytechnique, IP-Paris} \\
Palaiseau, France \\
jread@lix.polytechnique.fr}}

\maketitle

\begin{abstract}

Reinforcement Learning has drawn huge interest as a tool for solving optimal control problems. Solving a given problem (task or environment) involves converging towards an optimal policy. However, there might exist multiple optimal policies that can dramatically differ in their behaviour; for example, some may be faster than the others but at the expense of greater risk. We consider and study a distribution of optimal policies. We design a curiosity-augmented Metropolis algorithm (CAMEO), such that we can sample optimal policies, and such that these policies effectively adopt diverse behaviours, since this implies greater coverage of the different possible optimal policies. In experimental simulations we show that CAMEO indeed obtains policies that all solve classic control problems, and even in the challenging case of environments that provide sparse rewards. We further show that the different policies we sample present different risk profiles, corresponding to interesting practical applications in interpretability, and represents a first step towards learning the distribution of optimal policies itself. 

\end{abstract}


\section{Introduction}

Reinforcement Learning (RL) is a sequential decision framework where a model (agent) has to solve a task in multiple steps and find the optimal strategy (policy, $\pi$) to do so. The agent interacts directly with its environment and learns through this experience. After each time step $t$ that the agent performs an action $a_t$, the environment returns a new state $s_{t+1}$ and a reward $r_t$. The agent then seeks to maximise the expected return i.e. the discounted sum of rewards received over time: 
$G=\mathbb E[r_0+\gamma r_1 +\gamma^2 r_2 + \dots \gamma^t r_t + \dots |s_0]$; where gamma is a discount factor of future rewards. We define the value of state $s$, $V^\pi(s)$, as the expected return obtained when following policy $\pi$ from state $s$; and the Q-value $Q^\pi(s,a)$ as the state value obtained after performing action $a$ in state $s$. \newline
$V^{\pi}(s)  =\mathbb E_{\pi}\left[\sum_{k=0}^\infty \gamma^{k}r(s_{t+k+1})|s\right]=\sum_{a}\pi(a|s)Q^\pi(s,a)$

An optimal policy $\pi^*$ is a policy that reaches the maximal value possible for all states. Such a policy always exists but might not be unique \cite{MDP}. For instance, given policies $\pi_{\theta_i}$ parameterized by $\theta_i$, in parameter space $\Omega$, there are many $\theta_i$ that represent an optimal policy. Most of the literature focuses on finding models and algorithms that learn policies iteratively and converge towards a unique optimal policy \cite{Sutton}. But considering that many optimal policies exist, we propose a process to generate different optimal policies.  


Indeed, even-though all desired policies may be optimal, they may differ in their behaviour towards solving the task. We can argue that an optimal policy is the fastest and most consistent one that solves the given task. However, these two concepts can be antagonist depending on the policy's risk profile. Two policies can be optimal but one might take more risk than the other and solve the task faster. Sutton and Barto \cite{Sutton} give an example of policies obtained using Q-learning or SARSA and show that in the cliff problem (see table \ref{tab:Environments}), both methods solve the task but Q-learning takes more risk by walking near the edge while SARSA takes a longer, safer, path. 


As many optimal policies exist, then a distribution of optimal policies also exists. Learning such a distribution would allow to sample different optimal policies and choose the one which profile suits the best the task needs. For instance, the mode of the learnt distribution corresponds to the minimum variance optimal policy. However, to learn a distribution, many samples are needed. This paper addresses the question of whether or not it is possible to sample many different optimal policies. While all existing approaches focus on converging towards one optimal policy, we try to find a solution to generate many different ones that could correspond to different profiles. 


We adapted the Metropolis algorithm \cite{Metropolis} for RL in order to obtain a Markov chain of optimal policies whose stationary distribution corresponds to the distribution of optimal policies. We thus propose to sample a suitable distribution featuring {\it intrinsic} and {\it extrinsic} rewards (see Eq. \eqref{eq_fer}). We show that the obtained policies effectively adopt different behaviours while solving the desired tasks.

\section{Related work and brief introduction to Reinforcement Learning}


We cited Q-learning \cite{Q} and SARSA \cite{Sutton} above as classic methods used in RL. These two methods build Q-tables that store for every state-action pair the mean reward obtained afterwards until converging towards good estimations. After convergence, the optimal policy is the one that always follows the action that returns the highest value. While these methods try to evaluate Q values directly, other parameterize the policy itself like policy gradient methods \cite{PG}. The advantage of the latter being that these are capable to learn stochastic policies but also perform in continuous action spaces.


Deep Learning had a massive impact on RL and had been naturally incorporated into the cited approaches. This led to the emergence of Deep Q-learning \cite{DQN}, which uses deep neural networks to evaluate Q-values in large state spaces, or to Deep Policy Gradients that use a neural network parameters as policy parameters. Some other methods bridged the gap between these approaches like Actor Critic methods that use 2 neural networks, one that acts as policy and an other that evaluates state values \cite{Actor-critic}. 


Generative Adversarial Networks (GAN) \cite{GAN} are closely related to Actor Critic where the actor network can be assimilated to a generator and the critic to a discriminator \cite{GAC}. Indeed, GAN have been used in imitation learning for instance where expert demonstrations are used to learn a policy \cite{GAIL}. In this case the generator has to generate trajectories while the discriminator has to decide whether the trajectory\footnote{\textit{In RL literature, a trajectory is a successions of state action pairs or states alone. A trajectory can be longer than one episode. In this work, we constraint a trajectory to not exceed the length of an episode. Therefore, in this context, the words episode and trajectory are equivalent}} comes from the available demonstration or not. However instead of learning a new policy or outputting many different ones, the aim is to mimic the expert behaviour. Yet again, there is a close connection between this approach and GAN \cite{Finn}.


However these methods converge towards a unique optimal policy, rather than generate different optimal ones. All these advances show that using generative approaches to learn an optimal policy is possible but for our knowledge, there is yet no generative model that is used to generate different optimal policies that output different behaviours and different risk profiles. This work is a first step towards generating such samples in the hope of being then able to learn a distribution of optimal policies in future work. 

\section{Metropolis and Direct Policy Search}


Our main objective is to sample from the distribution of optimal policies. In this section we present a simple adaptation of Metropolis algorithm to RL problems. 

\subsection{Metropolis Algorithm}


Let $\theta$ denote the parameters of a policy $\pi_{\theta}$. Rather than searching directly for optimal $\theta$, several works consider sampling policy parameters $\theta_i$ using Markov Chain Monte Carlo algorithms \cite{BMCMC,TDMC}. For instance, in \cite{TDMC}, Hoffman et al.\ showed that for direct policy search, sampling directly from a trans-dimensional distribution that is proportional to the reward performs better than classic simulations methods. Therefore, we model a distribution $f(\theta)$, i.e the target distribution, that is proportional to the expectation of a monotonically increasing function $U$ of the empirical return, i.e., it should be maximal when the return is maximal; a utility function $U(\tau)$. 

\begin{equation}
    f(\theta)\propto p(\theta)\eta(\theta).
\end{equation}
where $p(\theta)$ is a prior over $\theta$ (in the simple case, uniform) and $\eta$ the performance of $\theta$ wrt the environment: 
 
 \begin{align}
    \label{eq:eta}
     \eta(\theta) =  \mathbb{E}_{p(\tau|\theta)}[U(\tau)] = \int U(\tau) p(\tau|\theta)d\tau,
 \end{align}
where $\tau$ denotes a trajectory (i.e. the succession of states visited), $p(\tau|\theta)$ is the distribution of $\tau$ (i.e., the one induced by following policy $\pi_\theta$), and $U(\tau)$ the utility of $\tau$.
Provided that $\eta(\theta)$ is non-negative, we can sample it via a Metropolis-Hastings (MH) algorithm. 

Denoting the empirical return by $\widetilde{G}(\tau) = \sum_{t=0}^{L-1}\gamma^t r_t$,
a usual choice is $U(\tau) = \exp(\widetilde{G}(\tau)/T)$. Therefore: 
\begin{equation}\label{eq_mi_tar}
    f(\theta;T) \propto p(\theta)\eta_T(\theta) = p(\theta)\mathbb{E}_{p(\tau|\theta)}[e^{\widetilde{G}(\tau)/T}]
\end{equation}
where $T$ denotes an inverse temperature parameter.  

In practice, in order to apply the MH algorithm on $f(\theta)$, the evaluation of $\eta(\theta)$ must be substituted with an unbiased estimate over $N$ episodes, 
\begin{align}\label{eq_luca}
\eta(\theta) &= \mathbb{E}_{p(\tau|\theta)}[U(\tau)] \nonumber\\
&\approx \bar{U}_N(\theta) =  \frac{1}{N}\sum_{i=1}^N  U(\tau_i), \quad \tau_i \sim p(\tau|\theta),
\end{align}
where $\bar{U}_N(\theta)$ denotes the empirical utility for $\theta$ over $N$ episodes $\tau_i$. 
Algorithm \ref{alg:naive} consider the noisy evaluation of the utility function in Eq. \eqref{eq_luca}, and is called {\it Monte Carlo-within-Metropolis} \cite{llorente2021survey,llorente2022optimality}. 

\begin{algorithm}
\caption{Monte Carlo-within-Metropolis for RL}\label{alg:naive}
\begin{algorithmic}
\Require $K$: the number of iterations, $N$: number of episodes
\State Initialise Agent $\pi$
\State $\theta_0 \sim \mathcal{N}(\mathbf{0},\sigma_p^2{\bf I}_D)$
\For{$k$ from $1$ to $K$}
    \State ${\theta'} \sim \mathcal{N}(\theta_{k-1},\sigma_p^2{\bf I}_D)$
    \State Run $N$ episodes with $\pi_{\theta_{k-1}}$ and compute  $\bar{U}_N(\theta_{k-1})$
    \State Run $N$ episodes with $\pi_{\theta'}$ and compute $\bar{U}_N(\theta')$
    \State $\beta \gets \dfrac{p(\theta')\bar{U}_N(\theta')}{p(\theta_{k-1})\bar{U}_N(\theta_{k-1})}$
    \State $\alpha \gets \min(1,\beta)$
    \State $\epsilon \sim \mathcal{U}_{[0,1]}$
    \If{$\epsilon<\alpha$}
        \State $\theta_k \gets \theta'$
    \Else
        \State $\theta_k \gets \theta_{k-1}$
    
    \EndIf
\EndFor

\State \Return $\{\theta_k\}_{k=1}^K$
\end{algorithmic}
\end{algorithm}

\section{Curiosity Augmented Metropolis}

\subsection{Prior Bias}

In section \ref{Simple results}, we showed that $\{\theta_k\}_{k=1}^K$ are highly correlated; due to a scaling factor among them. Indeed, as  we considered that $p(\theta)\sim\mathcal{U}[-a,a]$, there is no bound to $\theta_i$ values and therefore we obtain $\theta_{k+1} = \lambda\theta_k$, with $\lambda$ a constant. A solution is to constraint $\theta_k\in[-1,1]^D$ but it is not possible to use $p(\theta)\sim\mathcal{U}[-1,1]$, as it would cancel out anyway during the calculation of $\beta$ (see algorithm \ref{alg:naive}). However it is possible to measure the variance of $\theta_k$ from $[-1,1]$:
\begin{equation}
    \sigma^2_{\theta_k} = \frac{1}{D}\sum_{j=0}^D\mathbbm 1_{\theta_{kj}\notin[-1,1]}(\theta_{kj}^2-1)^2,
\end{equation}
and therefore we define: $p(\theta)=e^{-\sigma^2_{\theta}}$

\subsection{Trajectory bootstrap}

In Algorithm \ref{alg:naive}, we had to run $N$ episodes twice to estimate the mean return of $\theta_{k-1}$ and $\theta'$. We could store the return of $\theta_{k-1}$ and use it in subsequent runs. However, the method shown in Algorithm \ref{alg:naive} was preferred to penalise the possible variance of the returns obtained using $\theta_{k-1}$. 

An alternative to still penalise variance while avoiding to run the $N$ episodes twice is to make use of importance sampling and the advantage function \cite{Sutton}. Intuitively, it corresponds to the extra reward that the agent could obtain by taking action $a$ over a random action: $A_{\theta}(s,a)= Q_{\theta}(s,a)-V_{\theta}(s)$. More specifically, by defining $G$ as a function of $\theta$:
 \begin{equation}
     G(\theta)=\int \widetilde{G}(\tau)p(\tau|\theta)d\tau
 \end{equation}
 the \emph{advantage} of $\theta'$ over $\theta_{k-1}$:
\begin{equation}
    \begin{split}
        G(\theta') & = G(\theta_{k-1})+\mathbb E_{\theta'}\left[\sum_{t=0}\gamma^tA_{\theta'}(s_t,a_t)\right]\\
        & = G(\theta_{k-1})+\sum_s\rho_{\theta'}(s)\sum_a\pi_{\theta'}(a|s)A_{\theta'}(s_t,a_t)\\
        & \approx G(\theta_{k-1})+\sum_s\rho_{\theta_{k-1}(s)}\sum_a\pi_{\theta'}(a|s)A_{\theta'}(s_t,a_t)\\
        & = G(\theta_{k-1})+\mathbb E_{s\sim\rho(\theta_{k-1})}\left[\sum_a\pi_{\theta'}(a|s)A_{\theta'}(s_t,a_t)\right]\\
        & = \medmath{G(\theta_{k-1})+\mathbb E_{s\sim\rho(\theta_{k-1})}\mathbb E_{a\sim\pi_{\theta_{k-1}}}\left[\frac{\pi_{\theta'}(a|s)}{\pi_{\theta_{k-1}}(a|s)}A_{\theta'}(s_t,a_t)\right]}\\
        & \approx G(\theta_{k-1})\\
        & \quad +\medmath{\mathbb E_{s,a\sim\rho(\theta_{k-1}),\pi_{\theta_{k-1}}}\left[\Pi(r+\gamma(V_{\pi_{\theta'}}(s')-V_{\pi_{\theta_{k-1}}}(s))\right]},
    \end{split}
\end{equation}
where $\Pi=\frac{\pi_{\theta'}(a|s)}{\pi_{\theta_{k-1}}(a|s)}$. $\rho$ is the state visitation frequency and we used $\rho_{\theta'}(s) \approx \rho_{\theta_{k-1}}(s)$ because we already have trajectories sampled from $\pi_{\theta_{k-1}}$, so it is easier to obtain $\rho_{\theta_{k-1}}(s)$ than $\rho_{\theta'}(s)$. Finally, TD Error \cite{Sutton} was used as an estimator for the advantage in place of Q-values in order reduce variability. This way, the return obtained using $\theta'$ can be estimated using the behaviour of the agent parameterised by $\theta'$ on the trajectories obtained under $\theta_{k-1}$, noted $\pi_{\theta'}(\tau_{\theta_{k-1}})$. 

\subsection{Curiosity Module}\label{curiosity}

As shown in section \ref{Simple results}, the previous implementation fails on Griworld and Cliff environments. One of the main reason explaining this failure is that the rewards in these environments are sparse. This means that a reward of -1 at each time step does not bring much information about the state of the agent and how close it is from the goal. The only valuable information are obtained either on the pit or when reaching the goal. In a gradient free approach like Metropolis where there is no criteria to drive the search, it means that the agent should first reach the goal by luck before improving. Moreover, the fact that there is no valuable information gathered at each run makes the process get stuck at some points in parameter space as all $\theta'$ are drawn from $\mathcal{N}(\theta_{k},\sigma^2_p \textbf{I}_D)$.

It is therefore necessary to implement a mechanism that drives exploration dynamically. Inspired by the approach in \cite{ICM}, we propose a curiosity module that consists of a Neural Network that learns to predict the next state given the last state of an agent. This way, if the network learns to predict the trajectory of $\theta_k$ and $\theta'$ tries something new, it will fail and output a large prediction error, called the intrinsic reward. The intrinsic reward is then added to the extrinsic reward, i.e the reward returned by the environment:
\begin{equation}\label{eq_fer}
    R(\tau) = \mu \widetilde{G}(\tau)+(1-\mu)\mathcal{L}^{(k)}(\tau),
\end{equation}
with $\mathcal{L}^{(k)}$ the prediction error (i.e. the loss) at step $k$ and $\mu \in [0,1]$. 
The utility is thus computed on this modified reward
\begin{align}
    \widetilde{U}_N(\theta) = \frac{1}{N}\sum_{i=1}^N U(\tau_i) \quad \tau_i \sim p(\tau|\theta),
\end{align}
where, if we take $U$ to be an exponential as above, we have
\begin{align}
    U(\tau_i)  & = \exp\{\mu \widetilde{G}(\tau_i)  + (1-\mu)\mathcal{L}(\tau_i)\}.
\end{align}
The output $\theta_k$ are all stochastic optimal policies that fool the curiosity module and that explore different trajectories. Algorithm \ref{alg:CAMEO} details the procedure. 

\begin{algorithm}
\caption{CAMEO}\label{alg:CAMEO}
\begin{algorithmic}
\Require $K$: the number of iterations, $N$: number of episodes, $\mu$ a constant
\State Initialise Agent $\pi$ and $\Phi$ the curiosity neural network
\State $\theta_0 \sim \mathcal{N}(\mathbf{0},\sigma_p^2{\bf I}_D)$
\For{$k$ from $1$ to $K$}
    \State ${\theta'} \sim \mathcal{N}(\theta_{k-1},\sigma_p^2{\bf I}_D)$
    \State Run $N$ episodes with $\pi_{\theta_{k-1}}$ and store trajectories $\tau_{\theta_{k-1}}$
    \For{each trajectory $\tau_{\theta_{k-1}}^{(i)}$
    }   
        \State $\mathcal{L}(\tau_{\theta_{k-1}}^{(i)}) \gets MSE(\Phi(\tau_{\theta_{k-1}}^{(i)}),\tau_{\theta_{k-1}}^{(i)})$
        \State $\mathcal{L}(\tau_{\theta'}^{(i)})\gets MSE(\Phi(\pi_{\theta'}(\tau_{\theta_{k-1}}^{(i)})),\pi_{\theta'}(\tau_{\theta_{k-1}}^{(i)}))$
        \State Train $\Phi$ using $\mathcal{L}(\tau_{\theta'}^{(i)})$
        \State $\mathcal{L}(\tau_{\theta_{k-1}})\gets \mathcal{L}(\tau_{\theta_{k-1}})+ \mathcal{L}(\tau_{\theta_{k-1}}^{(i)})$
        \State $\mathcal{L}(\tau_{\theta'})\gets \mathcal{L}(\tau_{\theta'}) + \mathcal{L}(\tau_{\theta'}^{(i)})$
        \EndFor
    \State Compute $\widetilde{G}(\tau_{\theta_{k-1}})$
    \State $\widetilde{G}(\tau_{\theta'}) \gets   \widetilde{G}(\pi_{\theta'}(\tau_{\theta_{k-1}}))$
    \State $R(\tau_{\theta_{k-1}}) \gets \mu \cdot \widetilde{G}(\tau_{\theta_{k-1}})  + (1-\mu) \cdot \mathcal{L}(\tau_{\theta_{k-1}})$
    \State $R(\tau_{\theta'}) \gets \mu \cdot \widetilde{G}(\tau_{\theta'}) + (1-\mu) \cdot \mathcal{L}(\tau_{\theta'}) $
    \State $\widetilde{U}(\theta_{k-1}) \gets  \dfrac{1}{N}\cdot\left(U(R(\tau_{\theta_{k-1}}))\right)$
    \State $\widetilde{U}(\theta') \gets \dfrac{1}{N}\cdot \left(U(R(\tau_{\theta'}))\right)$
    \State $\beta \gets \dfrac{p(\theta')\widetilde{U}_N(\theta')}{p(\theta_{k-1})\widetilde{U}_N(\theta_{k-1})}$
    \State $\alpha \gets \min(1,\beta)$
    \State $\epsilon \sim \mathcal{U}_{[0,1]}$
    \If {$\epsilon < \alpha$} $\theta_k \gets \theta'$
    \Else \quad $\theta_k \gets \theta_{k-1}$
    \EndIf
\EndFor
\State \Return $\{\theta_k\}_{k=1}^K$
\end{algorithmic}
\end{algorithm}

\section{Experiments and Results}

Our proposed framework has been tested in Classic Control Gym environments, Cartpole, Acrobot and cliff \cite{Gym} as well as on a gridworld. Table \ref{tab:Environments} shows environments details. 

\begin{table*}[h!]
    \centering
 \begin{adjustbox}{width=2\columnwidth, center}
 \begin{tabular}{||c || c || c || c || c || c || c || c || c || c ||} 
 \hline
 \multicolumn{1}{||c|}{\textbf{Environments}} &
 \multicolumn{1}{||c|}{Cliff} &
 \multicolumn{1}{||c|}{Cartpole} &
 \multicolumn{1}{||c|}{\begin{tabular}{@{}c@{}}Acrobot\end{tabular}} &
 \multicolumn{1}{c||}{Gridworld} \\
 \hline
 \multicolumn{1}{||c|}{\textbf{Snapshot}} &
 \multicolumn{1}{||c|}{\raisebox{-\totalheight}{\includegraphics[width=40mm, height=10mm]{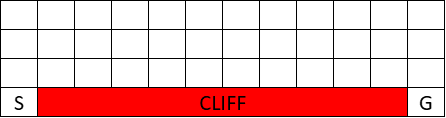}}} &
 \multicolumn{1}{||c|}{\raisebox{-\totalheight}{\includegraphics[width=20mm, height=15mm]{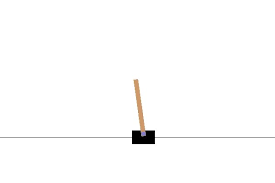}}} &
 \multicolumn{1}{||c|}{\raisebox{-\totalheight}{\includegraphics[width=20mm, height=15mm]{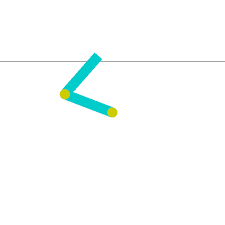}}} &
 \multicolumn{1}{c||}{\raisebox{-\totalheight}{\includegraphics[width=15mm, height=15mm]{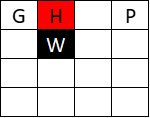}}} \\
 \hline
 \multicolumn{1}{||c|}{\textbf{State Space} $s_t \in$} &
 \multicolumn{1}{||c|}{$\{1,\ldots,48\}$}&
 \multicolumn{1}{||c|}{$\mathbb R^4$} &
 \multicolumn{1}{||c|}{$[-1,1]^4\times[-4\pi,4\pi]\times[-9\pi,9\pi]$} &
 \multicolumn{1}{c||}{$\{1,\ldots,16\}$} \\
 \hline
 \multicolumn{1}{||c|}{\textbf{Action Space} $a_t \in$} &
 \multicolumn{1}{||c|}{$\{0,1,2,3\}$}&
 \multicolumn{1}{||c|}{$\{0,1\}$} &
 \multicolumn{1}{||c|}{$\{0,1,2\}$} &
 \multicolumn{1}{c||}{$\{0,1,2,3\}$} \\
 \hline
 \multicolumn{1}{||c|}{\textbf{Reward} $r_t=$} &
 \multicolumn{1}{||c|}{-1 per move, 10 for the goal and -10 for the pit} &
 \multicolumn{1}{||c|}{+1 per time step} &
 \multicolumn{1}{||c|}{-1 per time step} &
 \multicolumn{1}{c||}{-1 per move, 10 for the goal and -10 for the pit} \\
 \hline
\end{tabular}
\end{adjustbox}
\caption{Specifications of Environments}
\label{tab:Environments}
\end{table*}

\subsection{Simple Metropolis implementation}\label{Simple results}

In this section we present the results obtained using the simple implementation detailed in algorithm \ref{alg:naive}. The agent is a neural network composed of 1 hidden layer of 8 neurons and a ReLU activation function. The average return was estimated on 20 episodes for every $\theta_i$. The metropolis algorithm was done on 200 timesteps for Cartpole and 500 hundred for Acrobot. 

\begin{figure}[h!]
    \centering
    \includegraphics[width=\columnwidth]{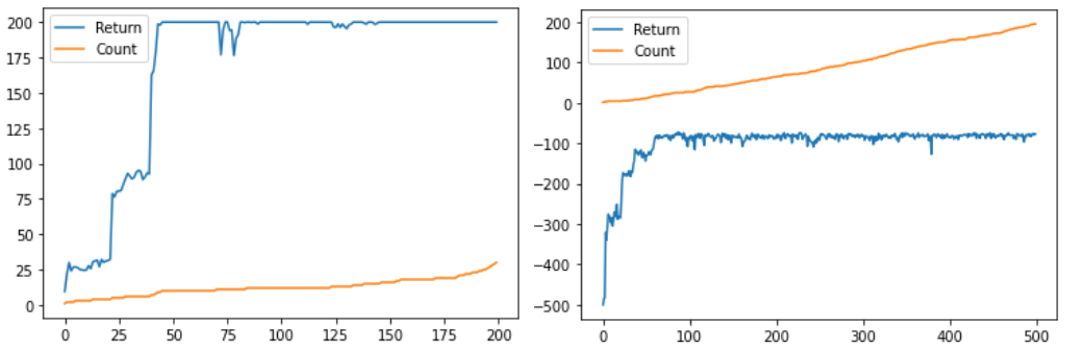}
	\caption{Average return for every new $\theta$ (in blue) and incremental count of the number of $\theta_i$ retained (in orange) on Cartpole (left) and Acrobot (right) using simple implementation.}
    \label{simple}
\end{figure}

Figure \ref{simple} presents the results obtained on Cartpole and Acrobot environments. Cartpole is solved while we converge towards a mean average return of -80 which is on par with great implementations according to gym leaderboard. Best implementation achieves a score of -40; we believe that our implementation can reach this score with enough iterations, which is the main drawback of MCMC methods. 

\begin{figure}[h!]
    \centering
    \includegraphics[width=\columnwidth]{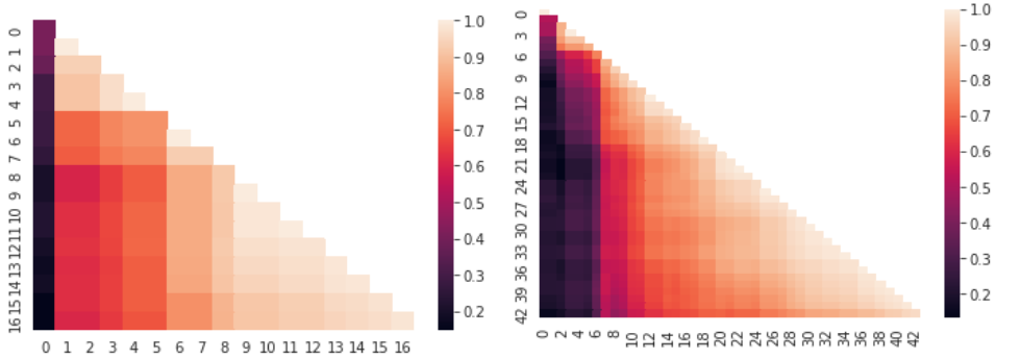}
	\caption{Cosine similarity between pairs of retained $\theta_i$ on Cartpole (left) and Acrobot (right) using simple implementation.}
    \label{simple_cosine}
\end{figure}

When visualizing the succeeding agents on Cartpole and Acrobot, it is not obvious if they effectively adopt different behaviours. We therefore plot the cosine similarity between all pairs of retained $\theta_i$ in figure \ref{simple_cosine}. It appears that succeeding $\theta_i$ are heavily correlated. 

However, the simple approach fails when confronted to Gridworld or Cliff. In both cases, the agent remains stuck, always performing the same action. The reasons for this failure are explained and tackled in section \ref{curiosity}. 

\subsection{Results for CAMEO}

\begin{figure}[h!]
    \centering
    \includegraphics[width=\columnwidth]{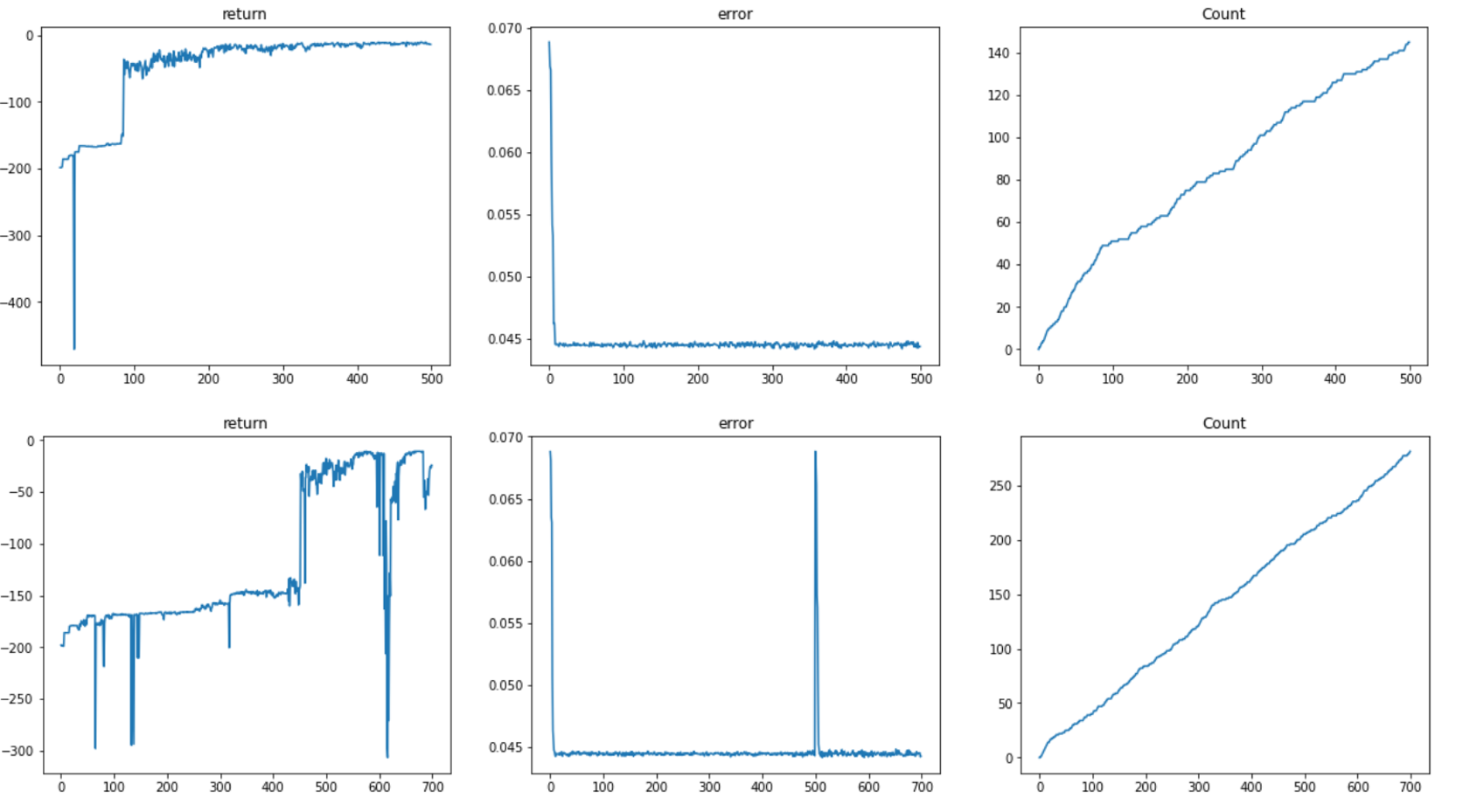}
	\caption{CAMEO Results on Cliff (above) and Gridworld (below). The figure presents the mean return, the Prediction error and the count of $\theta_i$ retained over time steps}
    \label{Cameo_results}
\end{figure}

In this section we present the results obtained for CAMEO on Gridworld and Cliff. We recall that for these environments, the rewards are sparse, and therefore the prior adaptation, as well as the curiosity module are necessary. Here we kept the same architecture for the agent while the curiosity module is a neural network of 1 hidden layer of size 150 with a ReLU activation function. As shown in figure \ref{Cameo_results}, the model succeeds in solving those two environments. We do not present the results on Cartpole and Acrobot as they are similar to those obtained without curiosity augmentation. 

\begin{figure}[h!]
    \centering
    \includegraphics[width=\columnwidth]{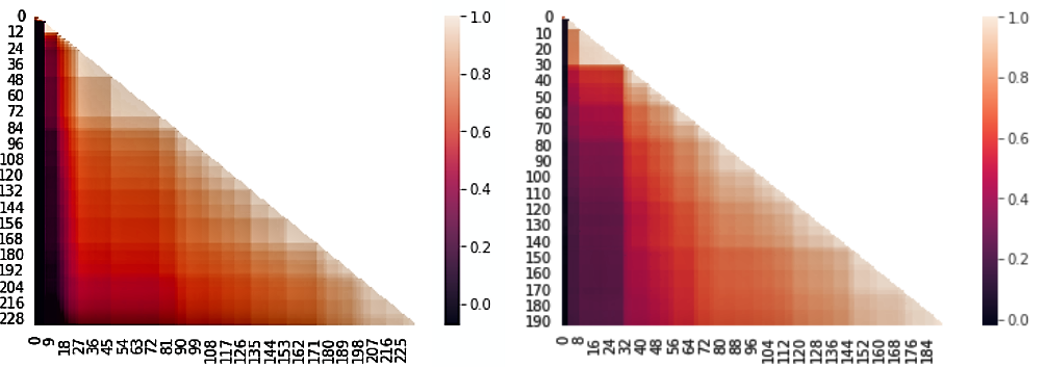}
	\caption{Cosine similarity between pairs of retained $\theta_i$ on Gridworld (left) and Cliff (right) using CAMEO implementation.}
    \label{Cameo_heat}
\end{figure}

\begin{figure}[h!]
    \centering
    \includegraphics[width=\columnwidth]{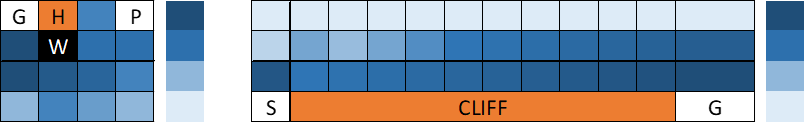}
	\caption{State visitation frequency aggregated on 100 policies obtained using CAMEO on Gridworld and Cliff. Less visited states are in light blue and most visited ones in dark shade}
    \label{frequency}
\end{figure}

It is interesting to note in figure \ref{Cameo_results} that the prediction error peaks when the model tries something new, even if it is not a good move but also that it recovers quickly afterwards. Moreover, we can note that the rate of $\theta_i$ retained is nearly linear, which shows that the ratio of rejection is low and therefore that the sampling process is efficient. Finally, figure \ref{Cameo_heat} shows that the $\theta_i$ retained are less correlated than in the simple implementation, which suggests that the curiosity module and the prior are effective. However non correlated weights do not always imply a different behaviour. Figure \ref{frequency} shows the aggregated state visitation frequency of 100 different policies that solve the problems. The most efficient (shortest) paths are the most taken but the state visitation frequencies are non negligible for other paths. Therefore, the learned policies effectively correspond to different behaviours. 

\section{Conclusion}

In the context of reinforcement learning, we were able to sample optimal policies on the fly and showed that the resulting behaviours are diverse. We also showed that our approach succeeded to output optimal policies even when the rewards structure of our problems were sparse, by using a curiosity module that encouraged exploration dynamically. 

Our approach still bears some limitations as the policy spaces of studied environments are discrete. As is generally the case with Monte Carlo methods, convergence is more challenging in high dimensions. To mitigate this, it is possible to replace the standard proposal distribution with a policy guided proposal that draws educated samples. The new distribution could therefore explore larger spaces by focusing on areas of interest. This will be subject to future work. 

\bibliographystyle{plain}
\bibliography{Bibliography.bib}

\end{document}